\documentclass[sigconf, 10pt]{acmart}
\usepackage{graphicx} 

\title{\huge Tri-Cam: Practical Eye Gaze Tracking via Camera Network}

\author{Sikai Yang}
\email{syang126@ucmerced.edu}
\affiliation{%
  \institution{University of California, Merced}
  \streetaddress{5200 North Lake Rd.}
  \city{Merced}
  \state{CA}
  \country{USA}
  \postcode{95343}
}

\author{Wan Du}
\email{wdu3@ucmerced.edu}
\affiliation{%
  \institution{University of California, Merced}
  \streetaddress{5200 North Lake Rd.}
  \city{Merced}
  \state{CA}
  \country{USA}
  \postcode{95343}
}

\usepackage{eucal}
\usepackage{bm}
\usepackage{makecell}
\usepackage{algorithm}
\usepackage{lineno}
\usepackage{algpseudocode}
\usepackage{amsmath}
\usepackage{subfigure}
\usepackage{booktabs}
\usepackage{multirow}
\usepackage{enumitem}
\usepackage{flowchart}
\usepackage{float}
\usepackage{times}
\usepackage{url}
\usepackage{amsfonts}

\usepackage{amssymb}
\usepackage{xspace}
\usepackage{graphicx}
\usepackage{tabu}
\usepackage{diagbox}
\usepackage{gensymb}
\usepackage{caption}
\usepackage{subcaption}
\usepackage{xcolor}

\renewcommand\footnotetextcopyrightpermission[1]{} 

\begin{document}
    \maketitle
    \pagestyle{plain}
    \section*{Abstract}

As human eyes serve as conduits of rich information, unveiling emotions, intentions, and even aspects of an individual's health and overall well-being, gaze tracking also enables various human-computer interaction applications, as well as insights in psychological and medical research.
However, existing gaze tracking solutions fall short at handling free user movement, and also require laborious user effort in system calibration.
We introduce Tri-Cam, a practical deep learning-based gaze tracking system using three affordable RGB webcams.
It features a split network structure for efficient training, as well as designated network designs to handle the separated gaze tracking tasks.
Tri-Cam is also equipped with an implicit calibration module, which makes use of mouse click opportunities to reduce calibration overhead on the user's end. 
We evaluate Tri-Cam against Tobii, the state-of-the-art commercial eye tracker, achieving comparable accuracy, while supporting a wider free movement area. 
In conclusion, Tri-Cam provides a user-friendly, affordable, and robust gaze tracking solution that could practically enable various applications.
    \section{introduction}
Gaze tracking refers to estimating where the user is looking.
Aside from existing human-computer interaction methods, such as gesture recognition~\cite{ren2021winect, chen2021rf, zhang2019towards, ma2021location} and haptic interaction \cite{10.1145/3544548.3580735}, gaze tracking brought new possibility of using human eyes as a new way of interaction \cite{VR, 10.1145/3379157.3391421, 10.1145/3448018.3458008}.
As eyes convey a wealth of information about emotions, intentions, and health, eye gaze tracking enables various applications in smart interaction \cite{duchowski2007eye, majaranta2014eye, salvucci2000identifying, borji2014defending, holmqvist2011eye, huang2018gazetouchpin, jansen2014eyescout,bhatti2021eyelogin}, psychology \cite{liversedge2000saccadic, hessels2019eye, holmqvist2017eye}, and medicine \cite{borji2014defending, wu2008detecting, pfeiffer2013gaze, vrij2000detecting}.



In this paper, we target the monitor-based gaze tracking, namely to infer where the user is looking on the screen.
It is a kind of remote gaze tracking, where the cameras are stationary, while the user is moving.
It has tremendous application market as people use computers and look at screens everyday.

\textit{Free movement and depth perception:}
Remote gaze tracking poses a great challenge to the interaction systems, due to free movement, namely, the relative location between the user's eyes and the cameras changes.
It necessitates depth perception capability for the gaze tracking systems to support practical and user-friendly gaze tracking applications.
Single camera \cite{ASgaze, huynh2021imon} cannot effectively handle depth, consequently demanding minimal user movement.
Although facial features may support depth perception \cite{kumar2013face}, in gaze tracking applications, users may frequently turn their head to look in various directions, making the solution less reliable.
Several works propose to use depth camera to handle free movement \cite{sun2015real, 10.1145/2638728.2641694}.
The Tobii eye trackers \cite{tobii} use infrared sensors to capture unique reflection from human eyes, and can be robust to distance.
However, these hardwares are relatively expensive.
A practical solution is to deploy multiple inexpensive non-depth cameras \cite{DVgaze, invisable, Hisadome_2024_WACV, gideon2022unsupervised}.
It leverages geometric relationship between the cameras to handle depth information, similar to the reason that human has two eyes.
However, too many cameras maybe also pose affordability issue \cite{hisadome2024rotation}.

In this work, we develop a practical gaze tracking system, Tri-Cam.
It utilizes three non-depth RGB webcams (each costing \$10) to handle free movement.
Compared to double-camera systems \cite{DVgaze,gideon2022unsupervised}, triple-camera enables an exclusive intra-validation mechanism via auxiliary multitasking.
It further exploits geometrical relationships between cameras to enhance gaze tracking accuracy.




\textit{Split gaze tracking neural network:}
The key to handling free movement lies in understanding the geometry between cameras, eyes, and the screen.
We analyze the gaze tracking task and divide it into camera-eye geometry and eye-screen geometry.
Accordingly, we design a split neural network structure that handles each geometry separately.
This division allows each part of the network to focus on its designated task, resulting in a lightweight network.


\textit{Camera-eye geometry and intra-validation:}
The camera-eye geometry refers to the vector from cameras to eyes.
This information can be represented by the coordinate where the eye is detected within the camera view.
Multiple camera-eye geometries form triangles that enable depth perception and consequently handle free movement.
To further exploit the camera-eye geometry, we develop an intra-validation design via auxiliary multi-tasking.
\cite{liebel2018auxiliary, adhikari2023argosleep}
It masks one camera, and uses the other two cameras to infer the camera-eye geometry of the masked camera, resulting in extra supervision for the network training, and enhanced gaze tracking performance.

\textit{Eye-screen geometry and weighted fusion:}
The eye-screen geometry refers to the gaze ray from the eyes to the gaze point on the screen.
This information is hidden within the eye images.
We rely on convolutional neural networks to extract this information.
However, eye images may have quality fluctuations due to reflection, occlusion or eye blinking.
To make the system robust to fluctuations, we design a discriminator-based weighted fusion mechanism that prioritizes higher-quality camera channels among the three cameras and two eyes, helping the system focus on relevant information.

\textit{Low-effort re-calibration via implicit data collection:}
Re-calibration can be a heavy burden, demanding users to look at visual targets to collect aligned gaze data for training.
Several works discusses the feasibility of utilizing opportunities in passive human attention \cite{yang2021vgaze, yang2022continuous, 10.1145/3173574.3174198}, e.g., video saliency.
Sugano \textit{et al.} \cite{sugano2015appearance} proposed to use mouse clicks as aligned opportunities, but assumed that the user’s gaze is directed at the mouse cursor, without discussing the potential error within the opportunities.
Huang \textit{et al.} \cite{huang2016building} conducted insightful observation on the alignment error during mouse click, taking in consideration of the size of interaction target and temporal perspective.
However, they missed valuable context information that can be extracted from daily computer usage, e.g., application context and the spatial distribution of click error.
To that end, we develop an implicit calibration module that utilizes aligned opportunities during mouse clicks to collect gaze data without explicit user participation.
By studying daily computer usage patterns, we establish three criteria: application context, press-release duration, and spatial distribution.
These criteria help filter and refine the implicitly collected data, thereby reducing the need for explicit data collection that requires active user attention.
We also demonstrate the feasibility of combining implicit and explicit calibration, providing users with an option to balance between accuracy and effort.

We collected 84,000 gaze data samples from 21 participants, allowing them to move freely in front of the monitor to simulate real-life computer usage scenarios. 
Extensive experiments are conducted to evaluate the gaze tracking performance of Tri-Cam, primarily comparing it to the state-of-the-art commercial eye tracker, Tobii Pro Spark \cite{housholder2021evaluating, tobii}.
The experiment results indicate that at a distance of 50cm from the monitor, Tri-Cam achieves an average gaze inference error of 2.06cm, which is near the 1.95cm error of Tobii eye tracker, while Tri-Cam supports a wider range of free-movement.

In summary, we make the following contributions:
\begin{itemize}[leftmargin=*]
    \item We analyze and partition the gaze tracking task into camera-eye geometry and eye-screen geometry.
    Accordingly, we design a light weight split neural network structure to handle the separated tasks.

    \item For camera-eye geometry, we propose a novel intra-validation mechanism to exploit geometric relationships between cameras and enhance gaze tracking accuracy.
    
    \item For eye-screen geometry, we devise a weighted-fusion strategy to improve robustness to image quality fluctuations.

    \item We develop an implicit calibration module that leverages mouse click opportunities, reducing the data collection overhead for the user.
    Based on real-life computer usage observation, we establish three criteria to filter and refine the implicitly collected data samples.

    \item We collect free-movement gaze data from 21 users and conduct extensive experiments to evaluate Tri-Cam.
\end{itemize}

    \section{Scenario Setup}
We first introduce the application scenario of remote gaze tracking, including several important technical configurations.

As an overview, in the context of remote gaze tracking, a user sits in front of a monitor and looks at the screen.
Meanwhile, multiple cameras on the monitor capture the user's eye image, based on which, the gaze tracking system estimates the user's gaze point on the screen.

\subsection{Cameras}
The cameras are inexpensive (\$10 each) webcams without depth or infrared capabilities, with a resolution of 1920$\times$1080. 
We positioned three webcams on the top of the monitor, ensuring that their relative position and orientation relative to the screen remain fixed.
This camera-on-monitor setup is less likely to change compared to cameras placed on a desktop, which are more prone to accidental touch or movement.


For each monitor and its associated cameras, we train a dedicated model. 
The implicit re-calibration module would help reduce the overhead of training new models for different monitor-camera setups. 
If the monitor is moved while the cameras remain fixed, the existing model does not require retraining. 
However, if the cameras on the monitor are significantly re-positioned, the new setup should be treated as distinct, necessitating model re-training.

\subsection{Gaze Point}

The gaze point indicates where the user is looking, referring to the specific pixel on the screen of 1920$\times$1080 resolution. 
Ground truth for the gaze point can be obtained by displaying a visual target on the screen and instructing the user to focus on it. 
The coordinate of the center pixel of the target serves as the ground truth for the gaze point.

\begin{figure}[ht]
    \hfill
    \begin{minipage}[t]{0.99\linewidth}
        \centering
        \includegraphics[width=\textwidth]{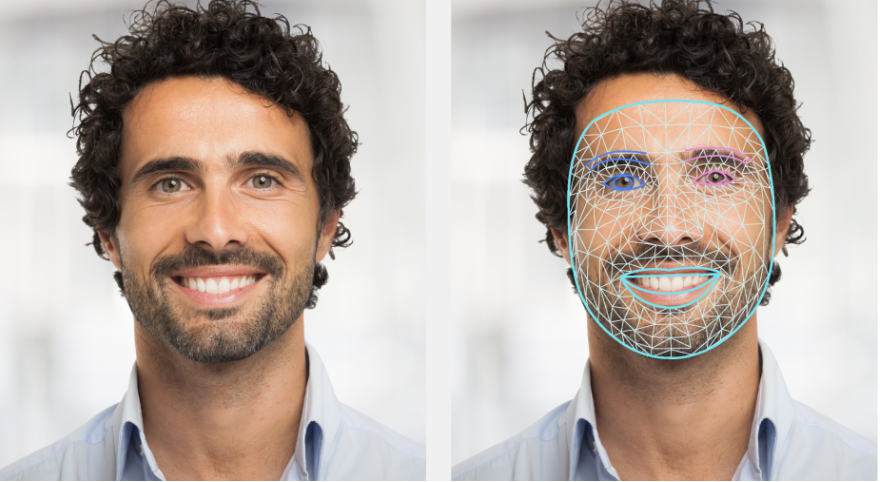}
         \vspace{-6mm}
        \caption{Eye detection via face detection}
        \label{fig_2_mediapipe}
        \vspace{-3mm}
    \end{minipage}
    \hfill
    \hfill
    \Description[bruh]{bruh bruh}
\end{figure}

\subsection{Eye Detection}

The system initially performs eye detection to determine if the user is looking at the screen and then locate the eyes within the camera's field of view. 
Several eye detection algorithms exist \cite{dxz_cv}, including OpenCV Haar cascades and Dlib eye detector. 
In our system, we utilize the MediaPipe eye detection algorithm \cite{mediapipe}.
Figure \ref{fig_2_mediapipe} (the image is sourced from the MediaPipe website) illustrates the workflow of the eye detection module. 
MediaPipe first detects the face and then identifies the eyes, providing the locations of the detected eyes within the camera's view. 
With this information, we can extract the eye images from the face image, resulting in four outputs: two cropped eye images and their corresponding coordinates within the camera's view.

\subsection{Practicality Standards}
We aim to develop a practical gaze tracking system that can be widely deployed and used with ease. 
Therefore, we identify three important standards for evaluating the practicality of such a system: affordability, robustness, and user-friendliness.

\subsubsection{Affordable Hardware}
First and foremost, a practical system should be affordable. Commercial eye trackers typically range from \$200 to \$4000, making them accessible primarily to customers with professional interests, such as gamers and streamers. 
For instance, the Tobii Pro Spark, which we used as a comparison in this work, costs \$3300. 
The gaming version, Tobii Tracker 5, is priced at \$250 but is limited to gaming assistance functions only. 
Both devices feature a central infrared camera and two infrared emitters on the sides. 
Depth cameras, on the other hand, are priced around \$100, making them prohibitively expensive for many users. 
Practical choices are often limited to non-depth RGB cameras, which are available for approximately \$10 to \$20.

\subsubsection{Robustness to Free Movement}
User movement will affect the spatial relationship between the eyes and the cameras, meaning that the eyes will appear at any position in the cameras' view. 
This variability can greatly impact the performance of eye gaze tracking.
In this paper, to ensure practicality and user experience, we allow users to move freely. 
This means that users can assume any comfortable body posture, and their eyes can be oriented in any direction relative to the cameras. 
They are only asked to maintain a healthy distance of 45 to 60 cm away from the screen.

Depth perception poses a significant challenge for gaze tracking systems with free movement.
Only with depth perception can the system project the gaze ray onto the screen to infer the location of gaze point.
Since depth cameras are not considered due to affordability reasons, we utilize multiple non-depth cameras to form a camera network and exploit their geometric relationships to perceive depth information.

\subsubsection{User-Friendliness}


Finally, a practical system should also minimize users' effort in preparation or redeployment. 
Gaze tracking systems typically require calibration before they are ready to use. 
Specifically, for learning-based gaze tracking, calibration is accomplished by training with the new user's data. 
To speed up training and reduce data collection overhead, the neural network size needs to be reduced to minimize the parameters to be trained. 
To achieve this, we design a split neural network structure (Section \ref{design 1}) that allows each part of the network to focus on its designated task with a limited network size. 
Additionally, we propose an implicit data collection module for Tri-Cam (Section \ref{design 4}) to further reduce data collection overhead and achieve low-effort redeployment for new users or new camera-monitor setups.
    \section{Design}

We now introduce the system design of Tri-Cam. 
We utilize three non-depth cameras to perceive depth information.
To jointly consider all cameras, in addition to the visual information hidden within human eye images, we rely on convolutional neural networks (CNN) to process eye images, and a multi-layer perceptron (MLP) to fuse all information.

We first analyze and divide the gaze tracking mechanism into two-fold: camera-eye geometry and eye-screen geometry.
Accordingly, we design a split neural network structure to handle the two geometries.
Then, to make full use of the multi-camera network, we design an intra-validation mechanism, which aims to make the model better capture the camera-eye geometry.
For eye-screen geometry, we perform weighted fusion for the eye images, using a joint discriminator that judges the importance of each captured eye image. 
Lastly, to reduce re-calibration (re-training) overhead, we design a low-effort implicit data collection module that makes use of cursor click opportunities.

\subsection{Gaze Tracking Geometry} \label{design 1}

In the context of gaze tracking, the user looks at the screen while the cameras focus on the user's eyes.
We divide the gaze tracking mechanism into two steps: 
camera-to-eye and eye-to-screen.

\subsubsection{Camera-Eye Geometry}
First, the cameras capture the user's image.
Then, via eye detection algorithm, the system locates user's eyes in each camera's view.
The eye center's pixel within the camera's view indicates the direction of the eye relative to the camera.
With multiple cameras and directions, and via triangulating, the system is capable of anticipating the position of the eye, or in another word, acquiring depth information.

\subsubsection{Eye-Screen Geometry}
As the eye position is acquired from the camera-eye geometry, based on that, the eye-screen geometry then infers the direction that the eye is looking in.
This information is embedded within the eye image, and can be learned by convolutional neural networks.
With this direction, and from the anticipated eye position, we can project the eye sight on to the screen to infer where the the user is looking at.

\begin{figure*}[ht]
    \hfill
    \begin{minipage}[t]{0.75\linewidth}
        \centering
        \includegraphics[width=\textwidth]{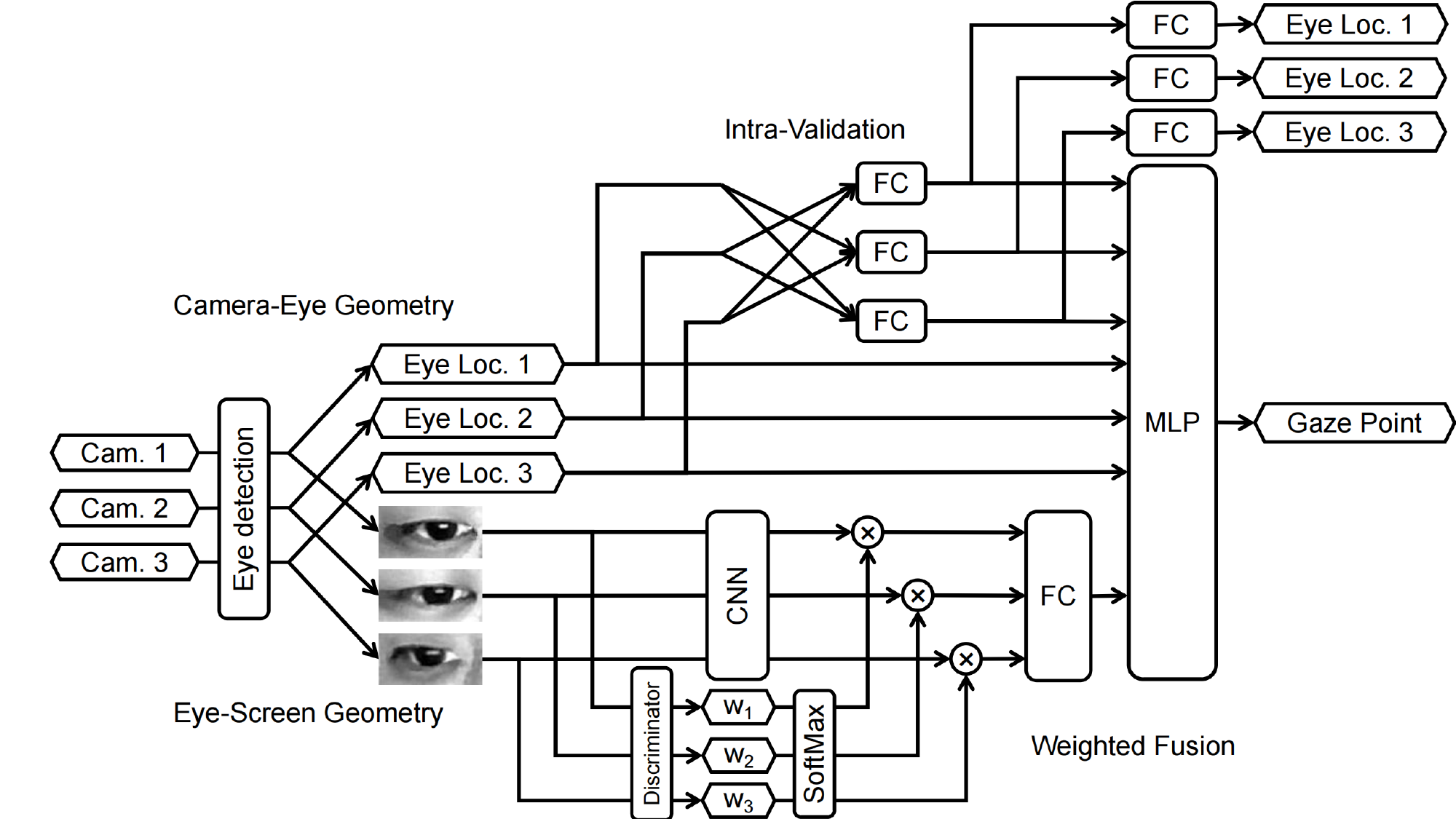}
         \vspace{-5mm}
        \caption{Tri-Cam Neural Network Architecture}
        \label{fig_1_main}
    \end{minipage}
    \hfill
    \hfill
    \Description[bruh]{bruh bruh}
\end{figure*}

\subsubsection{Split Neural Network Framework}
Based on the understanding of camera-eye geometry and eye-screen geometry, we design the gaze tracking neural network in a split form to separately handle the two geometries.
The network structure is shown in Figure \ref{fig_1_main}.
For simplicity, Figure \ref{fig_1_main} only shows the processing of the right eye.
The left eye will apply the same network structure, and share the final decision multi-layer perceptron to jointly output one gaze point.

In the upper part, we input the pixel-coordinates of the eyes within the camera's field of view, which are acquired via the eye detection algorithm. 
Specifically, if the eye is not detected in the camera's view at a certain moment, the eye location will be overwritten as -1, and the cropped image will be an all-black image when the eye is not detected.

In the lower part, we only feed it with the cropped eye image, as the eye sight information is entirely concentrated within this area.
The background image contains little useful information for the task of gaze tracking and should thus be discarded.
Motivated by a head-mounted eye gaze tracking work, Invisible-Eye \cite{invisable}, which uses low resolution but multiple images to achieve lightweight gaze direction, we also resize the resolution to 40$\times$20, which is still clear enough to distinguish the eye movement.
We use the six 40$\times$20-resolution eye images (2 eyes $\times$ 3 cameras = 6 images) to infer the eye-screen geometry, which is relatively lightweight, and thus guarantees fast calibration and re-training.

Via splitting the architecture into two parts, each part focus on its designated task.
However, since the lower part handles eye-screen geometry and processes images, which is large in dimension, we use a fully connected layer (FC) to further process its output and reduce its dimension to match the lower dimension of camera-eye geometry.
Finally, in the final decision multi-layer perceptron, both pieces of information are jointly considered to calculate the gaze point.

\subsection{Intra-Validation via Auxiliary Multitasking} \label{design 2}

To better master the camera-eye geometry, we design an intra-validation mechanism that leverages existing view angle information more effectively. 
This mechanism improves system accuracy through auxiliary multi-tasking.

\subsubsection{Design Motivation}

Assuming the gaze tracking system has been well calibrated and trained, based on the camera-eye geometries of two cameras, the system should be able to know the camera-eye geometry of the third camera.

Specifically, based on the coordinate of the eye in the view of two cameras, the system knows the directions from both cameras to the eye.
Based on these two directions, the system can deduce the position of the eye, and consequently determine the direction in which the third camera will observe the eye, i.e., the coordinate of the eye in the third camera.

A well trained gaze tracking system should be able to handle the relationship between the camera-eye geometries from three cameras.
Conversely, we expect to improve the system via guiding it to learn the relationship in the camera-eye geometries, via an auxiliary multitasking design \cite{liebel2018auxiliary}.

\begin{figure}[ht]
    \hfill
    \begin{minipage}[t]{0.75\linewidth}
        \centering
        \includegraphics[width=\textwidth]{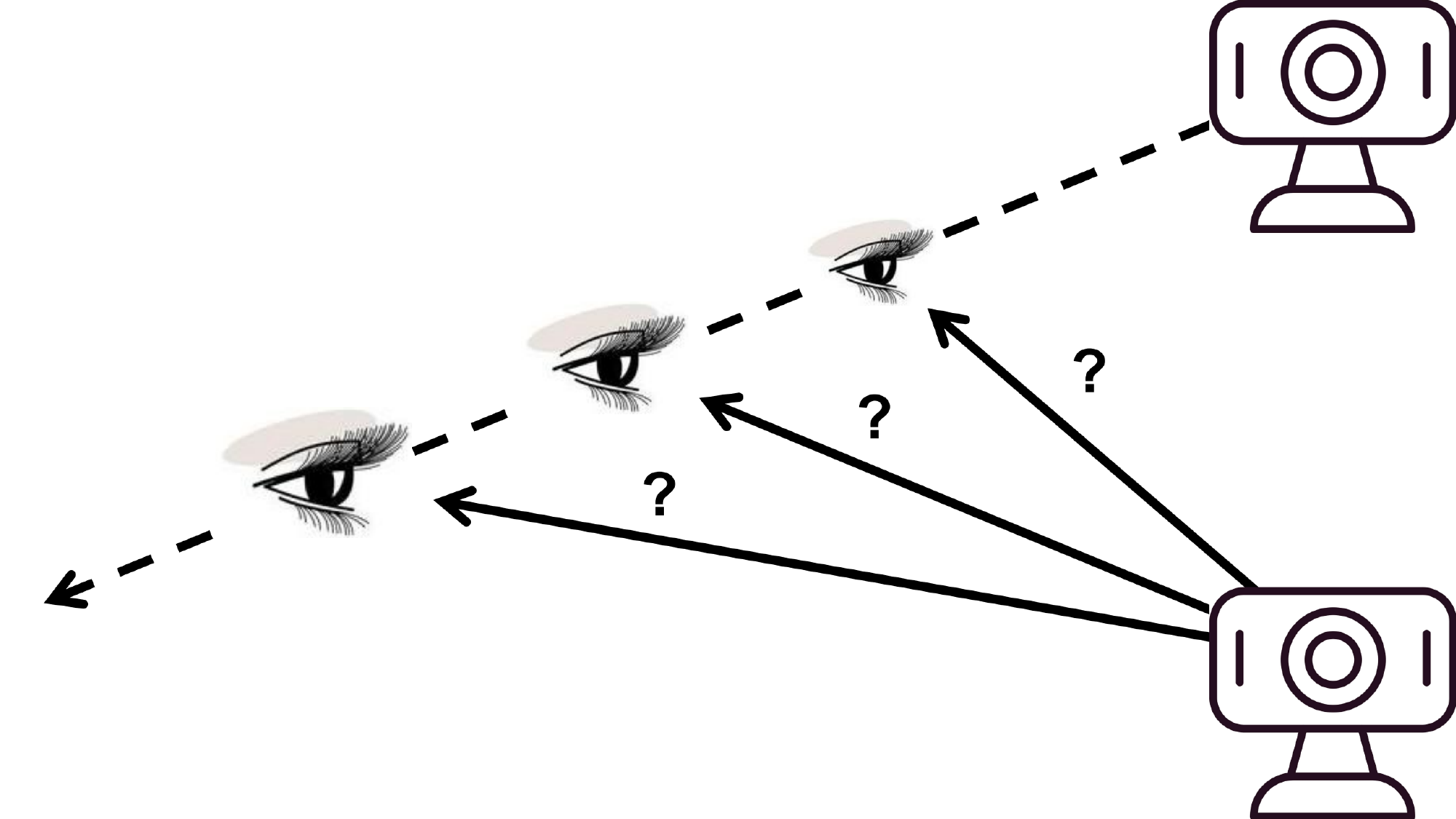}
        \vspace{-6mm}
        \caption{Two cameras are not enough to support the intra-validation mechanism}
        \label{fig_2_three}
        \vspace{-6mm}
    \end{minipage}
    \Description[bruh]{bruh bruh}
    \hfill
    \hfill
\end{figure}

\subsubsection{Intra-Validation Network Design}

Based on the above speculations, we fuse the camera-eye geometry information of camera 1 and 2, i.e., coordinate of eye in the camera's view, into fully connected layers and let it calculate the camera-eye geometry information of camera 3 as an auxiliary output.
We use the actual eye coordinate in camera 3's view as ground truth to calculate the auxiliary loss.
We also use camera 2 and 3 to predict the camera-eye geometry of camera 1, and use camera 1 and 3 to predict the camera-eye geometry of camera 2.
The auxiliary losses of the predicted camera-eye geometries will be used to perform back propagation to train the fully connected layers at the middle in conjunction with the back propagation of the main gaze point loss.
The auxiliary losses and the main loss will be combined into a joint loss.
We set a ratio of 0.1 for all three auxiliary losses as they are incorporated into the joint loss.
If the eye is not detected, the neural network pipeline will be multiplied by 0 before the last fully connected layer, so that the back propagation will not update the networks before it.

This architecture does not require extra information or measurement, but instead, masks and infers existing information.
The mechanism behind these auxiliary inferences is pertinent to the main task of the network, and thus could facilitate the training of related parts of the network.

\subsubsection{Why Three Cameras} \label{design 3}
The intra-validation design aims to harness the potential of the camera network, which is exclusive to systems with three or more cameras.
With one camera's information masked, two cameras remain, allowing them to form a triangle with their geometric information.

However, if there are only two cameras, it is not practical to infer the direction in which one camera will see the eye using the other camera. 
Webcams can only provide directional information. 
As a result, the possible location of the eye can vary along the direction, leading to uncertainty in the direction of the detected eye for the other camera, as depicted in Figure \ref{fig_2_three}.
Therefore it would not be sensible to conduct intra-validation between the two cameras, let alone expect an improvement in system performance.
While it may be feasible to measure the depth of an object with a non-depth camera using the size of the object in the image, in the task of gaze tracking, users' eye image size may often vary due to different eye openness and view angle, making it impractical to use the size of eye images to infer depth.


\subsection{Weighted Camera Fusion via Joint Discriminator}

The eye-screen geometry refers to the gaze ray from the eye to the gaze point.
It is embedded within the cropped eye image.
The six cropped eye images, derived from the combination of two eyes and three cameras, undergo processing by convolutional neural networks before being fused into the final multi-layer perceptron.

To enable the network to handle different image qualities, we design a weighted fusion mechanism. 
This mechanism utilizes a joint discriminator to assess the potential contribution of each cropped eye image and assigns lower weight to images with lower quality. 
This allows the network to prioritize valuable information for the final inference.

\subsubsection{Design Motivation}

\begin{figure} [ht]
    \hfill
    \subfigure[Blink]{\includegraphics[width=.23\linewidth]{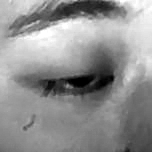}}
    \hfill
    \subfigure[Eye closed]{\includegraphics[width=.23\linewidth]{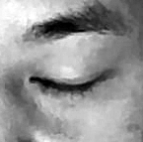}}
    \hfill
    \subfigure[Reflection]{\includegraphics[width=.23\linewidth]{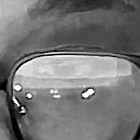}}
    \hfill
    \subfigure[Occlusion]{\includegraphics[width=.23\linewidth]{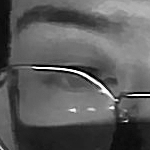}}
    \hfill
    \vspace{-3mm}
    \caption{Cropped eye image quality fluctuations}
    \label{fig_3}
    \vspace{-3mm}
    \Description[bruh]{bruh bruh}
\end{figure}

The cropped eye images may sometimes have low quality and do not provide very useful information.
For example, when the user blinks or temporarily close one of the eyes, as shown in Figure \ref{fig_3}, the eye image may be relatively less helpful for gaze tracking task.
In such cases, the system is expected to refer more to the image of the other eye.
Additionally, for users that wear glasses, the light reflected by the glass and the glasses frames may also occasionally block the eye image and degrade its quality, as shown in Figure \ref{fig_3}.
In such cases, the current camera happens to capture the reflection or the glasses frames occluding the eye, while other cameras may have better angle of view.

In conclusion, to handle fluctuation of eye image quality caused by both eye and camera factors, each cropped image should be individually judged before being fused into the final decision layer.
Images with lower quality and less important information should receive lower weights when being considered by the system.


\subsubsection{Weighted Fusion Network Design}

To that end, we design an eye image discriminator to judge the quality of the cropped eye images, and their potential importance to the gaze tracking task.
Based on the judgement, we conduct weighted fusion for the six information channels.
The lower part of Figure \ref{fig_1_main} show the structure of the weight fusion design.

Specifically, besides the main convolutional neural network, we also input the cropped eye images into the discriminator to assign a weight for each image ($w_1$, $w_2$, and $w_3$ in Figure \ref{fig_1_main}).
The weights will pass through a softmax layer to be jointly normalized, and then multiplied onto the main convolutional neural network output, so that images with lower quality or importance will be considered less in the final decision multi-layer perceptron.
The discriminator consists of two convolutional neural network layers and two fully connected layers, and outputs a 1-D scalar weight.

For simplicity, Figure \ref{fig_1_main} only depicts the processing of the right eye, while the cropped left eye images will also share the same discriminator to output the corresponding weights: $w_4$, $w_5$, and $w_6$.
Additionally, $w_1\sim w_6$ will all be fused into one softmax layer to output the jointly normalized weights.

\subsection{Low-Effort Re-calibration via Implicit Data Collection at Click Opportunities} \label{design 4}

\subsubsection{Design Motivation}

In the real application of gaze tracking, users are typically asked to calibrate the system before using it.
This is mainly due to biological difference, such as different eye sizes, shapes, and users' gaze habit.
The calibration process usually requests the user to look at a visual target rendered on the screen, meanwhile, the cameras capture the user's eye image, allowing the system to acquire an aligned data sample.
The aligned sample stores a valuable input-output pair.
The input stands for the eye image, and the output means the gaze point, i.e., the pixel coordinate of the target rendered on the screen.
Since the aligned data sample is captured while the user is staring at the target, this input-output pair can effectively supervise the calibration of the gaze tracking system.

However, such data collection requires active user attention, and thus can be laborious and expensive.
Furthermore, learning-based algorithms typically require heavy training data to optimize the neural networks.
In the worst scenario, the user may need to stare at target after target for a long time before the system is ready to use, which could cause eye fatigue and dissatisfaction.
To avoid that, we propose to relieve the requirement of explicit user attention, by identifying implicit alignment opportunities during daily computer use.

\subsubsection{Implicit Data Collection Module Design}
There are visual events that could attract or imply user's attention, e.g., video saliency \cite{yang2021vgaze,yang2022continuous}.
Similarly, we look for such events in mouse clicking \cite{sugano2015appearance,huang2016building}.
As a most common tool for human-computer interaction, electronic mouse controls the cursor that navigates within the screen.
The coordinate of the cursor on the screen can also be conveniently acquired.

When computer users click on buttons on the screen (e.g., the red-cross button that closes the window), they often use their sight to guide the cursor to navigate onto the button, and confirm its location before performing the click.
During the click, the cursor can be temporarily considered as a visual target.
As clicking is frequently iterated in daily computer using, it provides a great amount of aligned opportunities, where the user's gaze will align with, or approach the cursor.
Collecting data during these valuable opportunities does not explicitly require user's active attention, and thus could potentially reduce the calibration overhead.

However, it is also important to filter the implicitly collected data to guarantee data quality, as users may not look at the cursor during clicking in certain cases.
To examine the reliability of collecting aligned data during clicking, we recorded a total of 
6.89 
hours real-life active computer usage history from five users (user $\#16\sim\#21$), and summarize three major unaligned cases, as well as corresponding criteria that can help judge and refine the data collection quality.

Note that, we are now trying to evaluate whether the gaze point is aligned with the cursor's location, while the users are not explicitly instructed to look at anything, making it impossible to acquire the gaze point ground truth.
Therefore, we use the inference of Tobii Pro Spark eye tracker as the provisional ground truth of the gaze point to  to assess the extent of alignment between the cursor and the gaze point during clicking.
Furthermore, we only consider left clicks, as they are the primary mode of interaction using a mouse.

\begin{itemize}[leftmargin=*]
    \item Criterion A: active application context:

    The primary  factor that affects whether a user will look at the cursor during click is the application being used.
    In casual applications like file managing (file explorer, folder), internet browsing (Internet Explorer, Google Chrome, Fire Fox, not including online video web pages), and text editing (Notepad, Microsoft Doc., coding), gaze usually aligns well with the cursor when clicking.
    However, in applications such as games, the gaze point is almost never aligned with the cursor at all, as users are focusing on the visual feedback from the game.
    Additionally, many games hide the cursor during playing, particularly in first-person shooter games.
    Watching videos and movies also does not guarantee an alignment between the gaze and cursor, as users may randomly click anywhere on the video only to pause or resume playing, without looking at the cursor.

    Figure \ref{fig_implc_context} illustrates the average alignment error (the distance between the gaze point and the cursor location during clicking) across different active application contexts.
    According to the real-life computer usage record, gaming and watching videos are found to be entirely unreliable, whereas file managing (folder), internet browsing, and text editing still offer opportunities for implicit data collection.
    Therefore, we only select these three contexts as the scope of implicit data collection.
    Within these three contexts, we are able to filter out 15526 click opportunities, with an average alignment error of 4.90 cm.
    Building on this basic application context filtering, we then design criteria B and C to further improve the alignment accuracy.

\begin{figure}[ht]
    \hfill
    \begin{minipage}[t]{0.36\linewidth}
        \centering
        \includegraphics[width=\textwidth]{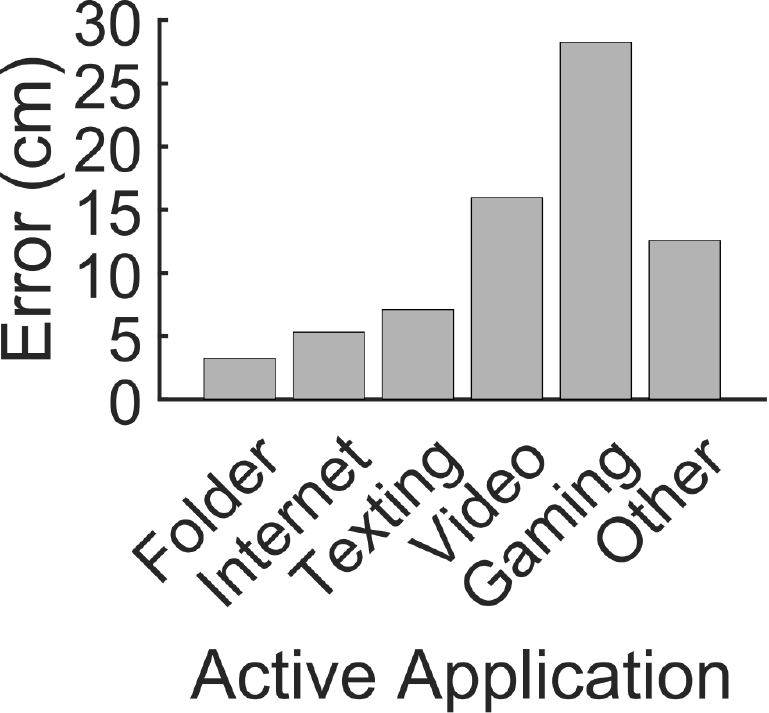}
         \vspace{-6mm}
        \caption{Error in different contexts}
        \label{fig_implc_context}
    \end{minipage}
    \hfill
    \begin{minipage}[t]{0.54\linewidth}
        \centering
        \includegraphics[width=\textwidth]{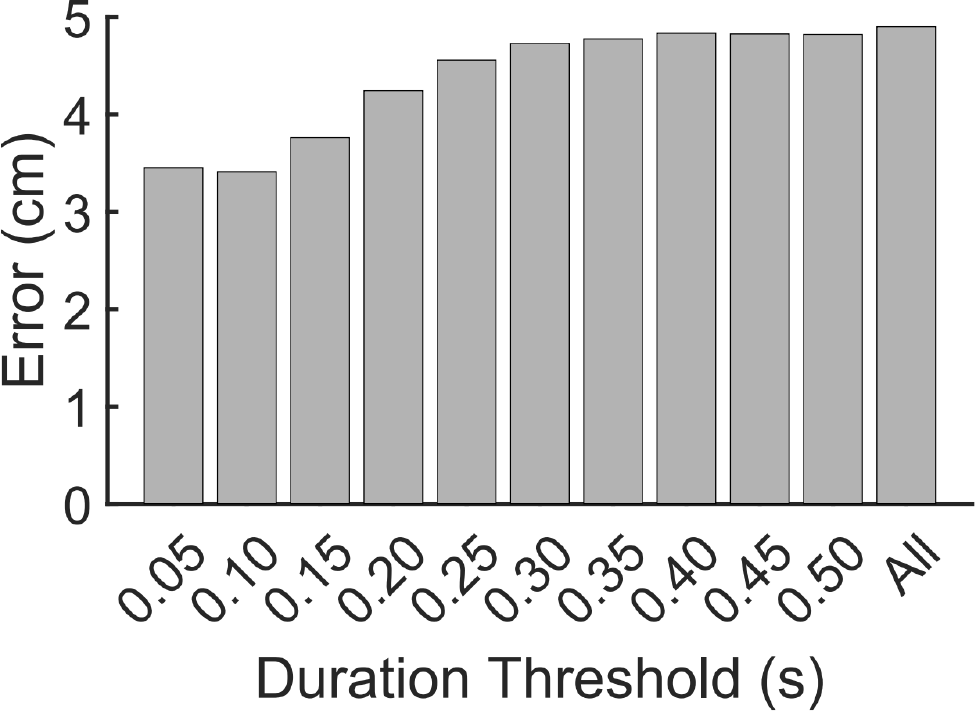}
         \vspace{-6mm}
        \caption{Error with different click duration threshold}
        \label{fig_implc_duration}
    \end{minipage}
    \vspace{-3mm}
    \hfill
    \hfill
    \Description[bruh]{bruh bruh}
\end{figure}

    \item Criterion B: press-release duration:

    Clicking involves both pressing and releasing actions on the mouse. 
    A click is identified when a release action immediately follows a press action. 
    However, a press-release action pair can also occur during dragging, where the duration between the press and release actions is longer.
    Users typically engage in dragging actions during file management (throwing a file into another window), and text editing (selecting a sentence).
    In these cases, users will only vaguely confirm the cursor's destination, and the duration between the press and release is long (e.g., $\>0.2s$).
    Longer duration implies higher likelihood of the user's attention being diverted to other contents before releasing the mouse (e.g., dragging scroll bar, while checking web contents).

    We filter the click data samples using different press-release duration thresholds, as illustrated in Figure \ref{fig_implc_duration}. 
    According to the real usage record, imposing a limited duration threshold helps reduce the average alignment error.
    With a duration threshold of 0.1 second, we filtered the 15,526 samples down to 2,434, resulting in an average alignment error of 3.41 cm. 
    However, further reducing the threshold to 0.05s does not necessarily reduce the error anymore, since users may occasionally click randomly with the mouse.
    As a result, we adopt the 0.1 second threshold as criterion B.

\begin{figure}[ht]
    \hfill
    \begin{minipage}[t]{0.98\linewidth}
        \centering
        \includegraphics[width=\textwidth]{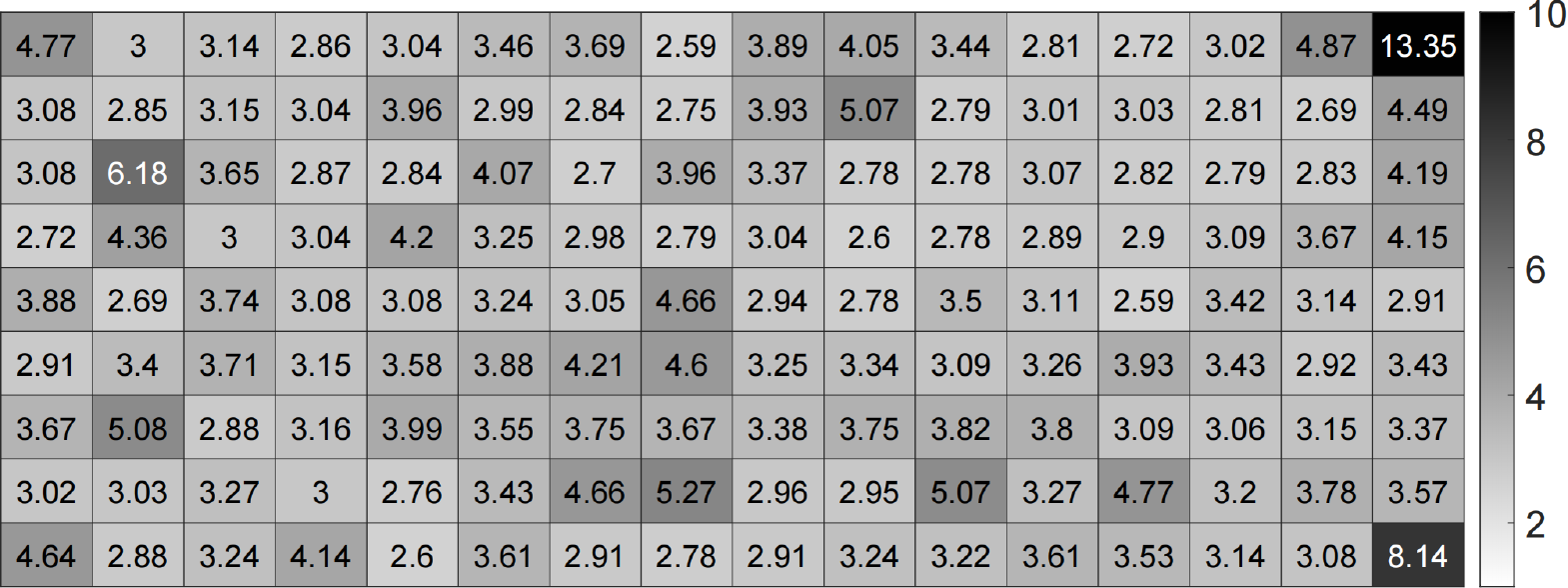}
         \vspace{-6mm}
        \caption{Spatial distribution of alignment error}
        \label{fig_implc_loction}
    \end{minipage}
    \hfill
    \vspace{-3mm}
    \Description[bruh]{bruh bruh}
\end{figure}

    \item Criterion C: click location:

    Last, we  examine the spatial distribution of the alignment error.
    Figure \ref{fig_implc_loction} shows the error distribution on the entire screen.
    The upper-right and bottom-right corners exhibit significantly higher errors, suggesting that users often do not precisely confirm the cursor location when clicking.
    
    As an explanation, skilled users may often heavily throw the cursor to the upper-right corner and click, as they know the close button is at that corner when the active window is maximized.
    The cursor being throttled to the upper-right corner will be confined by both the right and upper borders, and definitely land on the close button.
    Consequently, users only need to observe the window close as confirmation after they click the mouse, and do not need to look at the cursor.
    Similar rationale applies for the bottom-right corner, where the Windows system has a 'show-desktop' button that minimizes all windows and displays the desktop.
    However, this feature is less frequently used than the closed button.

    We use a Windows 11 system as the data collection platform, whose menu button is at the middle of the bottom taskbar.
    However, for Windows 10 and earlier versions, the menu button is positioned at the bottom-left corner, which may encounter similar issues if the system is to be deployed on these platforms.
    Additionally, Mac OS also features a menu button on the upper-left.

    We discard clicks that land within the range of 100 pixels from the four corner.
    As a result, 2,381 samples are filtered from the original 2,434 samples.
    The final average alignment error for the remaining 2,381 samples is 3.28cm.
    
\end{itemize}

With these criteria in place, we can effectively filter and refine the collected aligned clicks. 
During a recognized click opportunity, we capture the user's eye images via the camera networ. 
Along with recording the cursor location, we prepare an input-output pair as training data.

We save the captured eye images from the press moment till the release moment.
Specifically, the cameras are activated when the mouse is pressed and continue capturing images for 0.1 second or until the mouse is released.
According to criterion B, if the press-release duration exceeds 0.1 second, the click opportunity is recognized as invalid and discarded.
Since human has a short reaction delay \cite{gazzaniga2008cognitive, weinberg2010foundations, wickens2003engineering} after confirming the cursor location, we also record the eye images in the following 0.2 second after the release.
This yields extra 6 data samples (webcam video recording rate is 30Hz).

For the remaining 2381 click opportunities, the average press-release duration is 0.0786 second.
For each click opportunity, an average of 2.57 aligned data samples can be collected (e.g., a 0.01s duration can provide 1 sample at the press moment, and a 0.065s duration can only provide 2 samples, but 0.067s is able to provide 3 samples).
Additionally, with the 0.2s recording after the click, and using the 
6.89 
hours of active computer usage, we can implicitly collect 20,405 data samples for training.
The overall implicit data collection efficiency is 49.36 samples per minute.

To apply this module, the user simply operates the computer as usual, after setting up the cameras on the monitor.
The module will collect the click opportunities and train the neural network at the backend.
After a certain amount of time, the system will be ready to use.
If the user is not satisfied with the implicitly calibrated result, they can always revert to classical calibration, where the user is explicitly required to actively stare at visual targets to collect gaze data.

Combining implicit and explicit data collection methods is also a viable option. 
For example, a user initially train the network implicitly with 30 minutes of daily computer usage data but finds the result unsatisfactory.
In such cases, the user could spend an additional 3 minutes on explicit data collection, which originally would take 10 minutes of active staring that could cause eye fatigue, so that the implicit data collection module helped reducing the calibration effort.

    \section{Experiment}

We collect gaze data and conduct real experiment to examine the gaze tracking performance of Tri-Cam.
We mainly compare Tri-Cam to the state-of-the-art commercial eye tracker, Tobii Pro Spark, on various aspects.

\subsection{Experiment Setting}

\subsubsection{Baselines}

\begin{itemize}[leftmargin=*]

    \item Tri-Cam: 
    Our designed gaze tracking system, using three non-depth webcams.
    Tri-Cam uses cropped eye images only, which can be acquired from eye detection pre-processing.
    For post-processing, we conduct basic smoothing on the output of Tri-Cam to better fit the continuity of human gaze.
    
    \item Tobii Pro Spark eye tracker: 
    The state-of-the-art commercial eye gaze tracking product developed by Tobii Technology, who focuses on providing eye-tracking solutions for academic and commercial research. 
    Tobii Pro Spark features a a single integrated central infrared camera and two infrared transmitters on the sides.
    It is commonly used as gaze ground truth provider in other gaze tracking researches \cite{DVgaze, ASgaze}, due to its high accuracy and robustness.
    However, we are using Tobii Pro Spark as a baseline for comparison.
    
    \item DV-Gaze \cite{DVgaze}: 
    The latest deep learning-based gaze tracking work that uses Resnet\cite{resnet} and Transformer\cite{transformer}.
    DV-Gaze uses double camera to capture face image, instead of eye images, to infer gaze.
    For DV-Gaze, we use the two webcams at the two sides as its input.

    \item AsGaze \cite{ASgaze}:
    A geometry-based gaze tracking solution using a single non-depth RGB camera.
    AsGaze detects the iris boundary to estimate gaze ray.
    It can be configured to track gaze on different surface areas to suit various applications.
    We use the central camera as its input.
    However, its single-camera nature does not support free movement.
    Its error increases linearly as user moves away from the initial position.
    Therefore we only demonstrate its performance in the gaze angle section (Section \ref{sec_expm_angle}).
    
\end{itemize}

\begin{figure} [ht]
    \subfigure[\#1]{\includegraphics[width=0.125\linewidth]{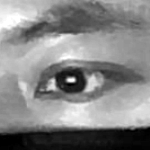}}
    \hfill
    \subfigure[\#2]{\includegraphics[width=0.125\linewidth]{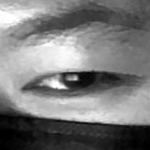}}
    \hfill
    \subfigure[\#3]{\includegraphics[width=0.125\linewidth]{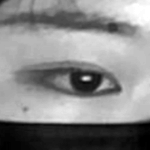}}
    \hfill
    \subfigure[\#4]{\includegraphics[width=0.125\linewidth]{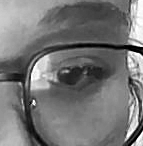}}
    \hfill
    \subfigure[\#5]{\includegraphics[width=0.125\linewidth]{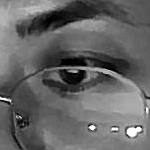}}
    \hfill
    \subfigure[\#6]{\includegraphics[width=0.125\linewidth]{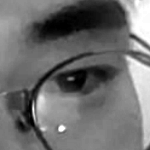}}
    \hfill
    \subfigure[\#7]{\includegraphics[width=0.125\linewidth]{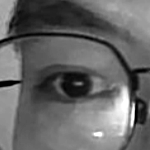}}
    \vspace{-3mm}
    
    \subfigure[\#8]{\includegraphics[width=0.125\linewidth]{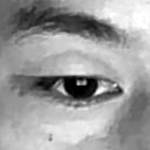}}
    \hfill
    \subfigure[\#9]{\includegraphics[width=0.125\linewidth]{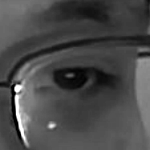}}
    \hfill
    \subfigure[\#10]{\includegraphics[width=0.125\linewidth]{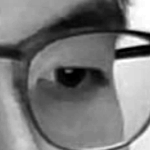}}
    \hfill
    \subfigure[\#11]{\includegraphics[width=0.125\linewidth]{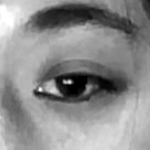}}
    \hfill
    \subfigure[\#12]{\includegraphics[width=0.125\linewidth]{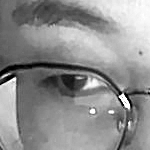}}
    \hfill
    \subfigure[\#13]{\includegraphics[width=0.125\linewidth]{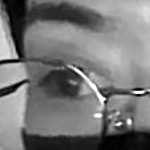}}
    \hfill
    \subfigure[\#14]{\includegraphics[width=0.125\linewidth]{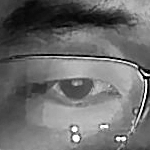}}
    \vspace{-3mm}
    
    \subfigure[\#15]{\includegraphics[width=0.125\linewidth]{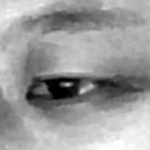}}
    \hfill
    \subfigure[\#16]{\includegraphics[width=0.125\linewidth]{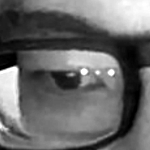}}
    \hfill
    \subfigure[\#17]{\includegraphics[width=0.125\linewidth]{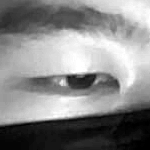}}
    \hfill
    \subfigure[\#18]{\includegraphics[width=0.125\linewidth]{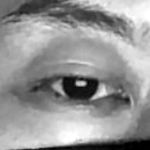}}
    \hfill
    \subfigure[\#19]{\includegraphics[width=0.125\linewidth]{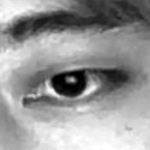}}
    \hfill
    \subfigure[\#20]{\includegraphics[width=0.125\linewidth]{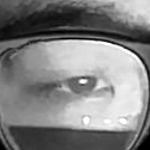}}
    \hfill
    \subfigure[\#21]{\includegraphics[width=0.125\linewidth]{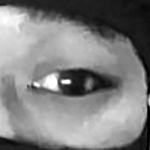}}
    \vspace{-5mm}
    \caption{All users' eye image}
    \vspace{-3mm}
    \label{fig_user_photo}
    \Description[bruh]{bruh bruh}
\end{figure}

\subsubsection{Data Collection}
Existing datasets do not specifically study the impact of free movement, i.e., changing pose, moving body, and turning head.
Additionally, they use Tobii as ground truth, which we are looking to compare with.
To build practical gaze tracking system that allows free movement, we collect new dataset from 21 users.
Among them, 11 wear glasses, 7 are female, and 14 are male.
For every user we collect 4000 aligned data samples, using visual targets as gaze point ground truth.
All users are allowed to blink, move body, and turn head.
They are only asked to maintain a healthy distance of 45$\sim$60cm away from the monitor.

Three cameras are evenly deployed on the top of the monitor, facing front to capture eye images, while the Tobii Pro Spark eye tracker will simultaneously infer the gaze point.

Before each data collection session, we always ask the user to calibrate the Tobii Pro Spark using its standard calibration procedure provided by Tobii Tracker Manager software.


The time limit for each data collection session is 30 seconds, after which the user will be asked to take a break for a least one minute to ensure precise gazing.
Sessions will be repeated till 4000 samples are collected.

\subsubsection{Evaluation Metric}

The gaze tracking error is defined as the Cartesian distance between the estimated gaze point and the visual target, both of which refer to a specific pixel locations on the monitor.
We convert the pixel distance to centimeters using the dimensions and resolution information of the monitor.

\begin{figure}[ht]
    \hfill
    \begin{minipage}[t]{0.85\linewidth}
        \centering
        \includegraphics[width=\textwidth]{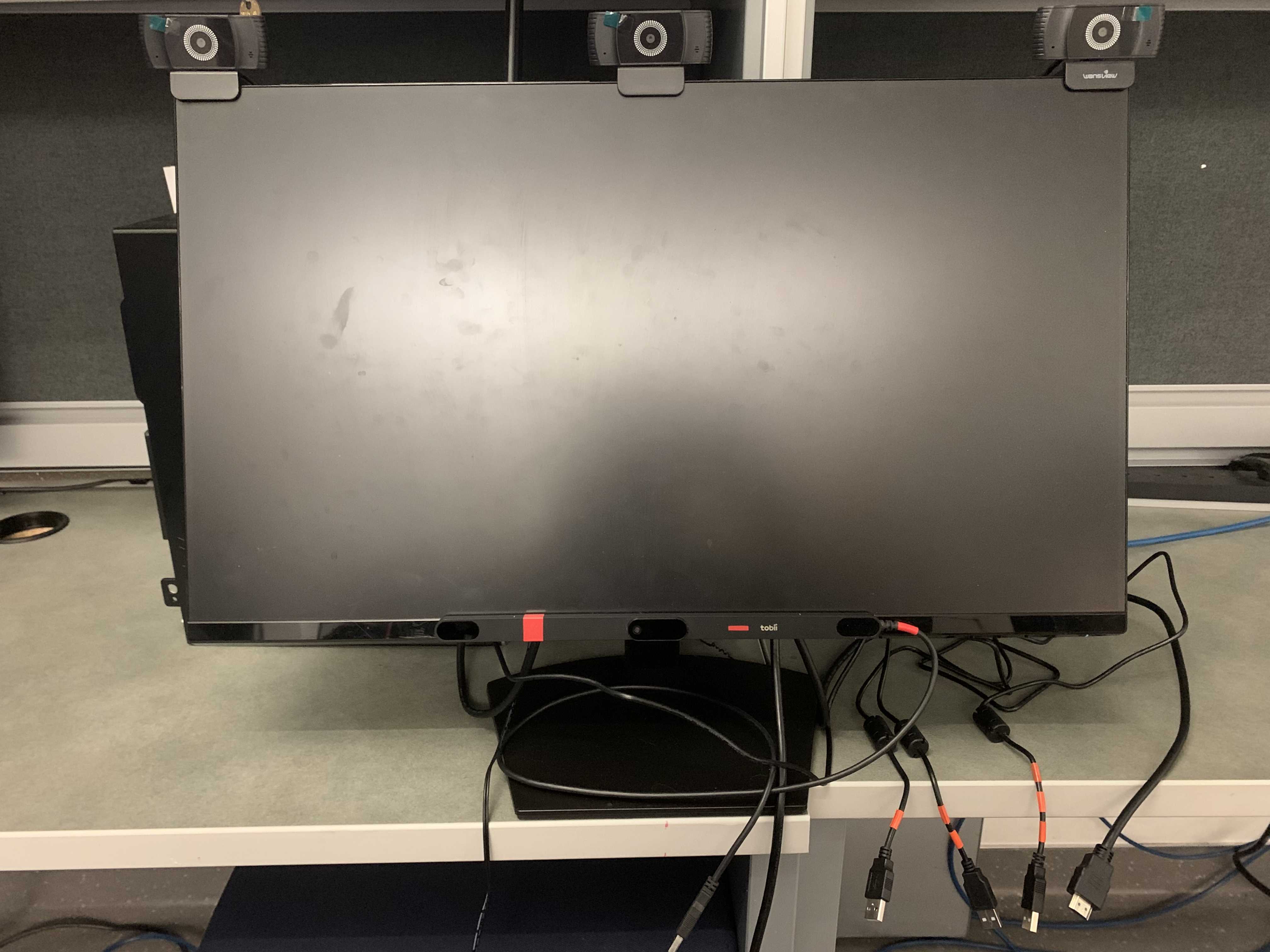}
         \vspace{-5mm}
        \caption{Monitor, webcams, and Tobii Pro Spark}
        \label{fig_expm_hardware}
        \vspace{-3mm}
    \end{minipage}
    \hfill    
    \hfill
    \Description[bruh]{bruh bruh}
\end{figure}

\subsubsection{Hardware}

As shown in Figure \ref{fig_expm_hardware}, we use the 27'' Dell SE2719H as the main monitor for experiment.
Its resolution is $1920\times1080$, and dimensions are $59.789\times 33.631cm$.
The Tobii Pro Spark Eye Tracker is installed at the bottom of the screen, facing front.
We use three Svarog webcams as camera input, whose video resolution is $1920\times1080$, and recording rate is 30Hz.
They are evenly placed at the top of the Dell monitor, facing forward.
For neural network model training, we use an Alienware M18 laptop, which is equipped with a NVIDIA® GeForce RTX™ 4090, 16 GB GDDR6 graphic card, and 24-Core Intel® Core™i9 13980HX Processor.
We also put the webcams on the Alienware M18 laptop monitor, and use it as a second screen for experiment.
Its resolution $1920\times1200$, and dimensions are $38.776\times 24.235cm$.

\subsection{Overall Performance}

The overall gaze estimation error for Tri-Cam, Tobii, and DV-Gaze are 2.0558cm, 1.9455cm, and 3.7077cm, respectively. 
Although Tri-Cam has higher error than Tobii, it reaches the similar level of accuracy using three inexpensive webcams.

\subsubsection{Different Users}

As personal computers are usually owned by individuals, a gaze tracking system should also be trained using personal gaze data, since it will also be mainly used by the individual.
Therefore, for user $\#1\sim\#16$, we train a model for each user individually, and test the gaze tracking accuracy on the user's own test data.
Specifically, we allocate 70\% of the data (2,800 samples) for training, 10\% (400 samples) for validation to select the best model during training, and 20\% (800 samples) as unseen data for testing.

\begin{figure}[ht]
    \hfill
    \begin{minipage}[t]{0.99\linewidth}
        \centering
        \includegraphics[width=\textwidth]{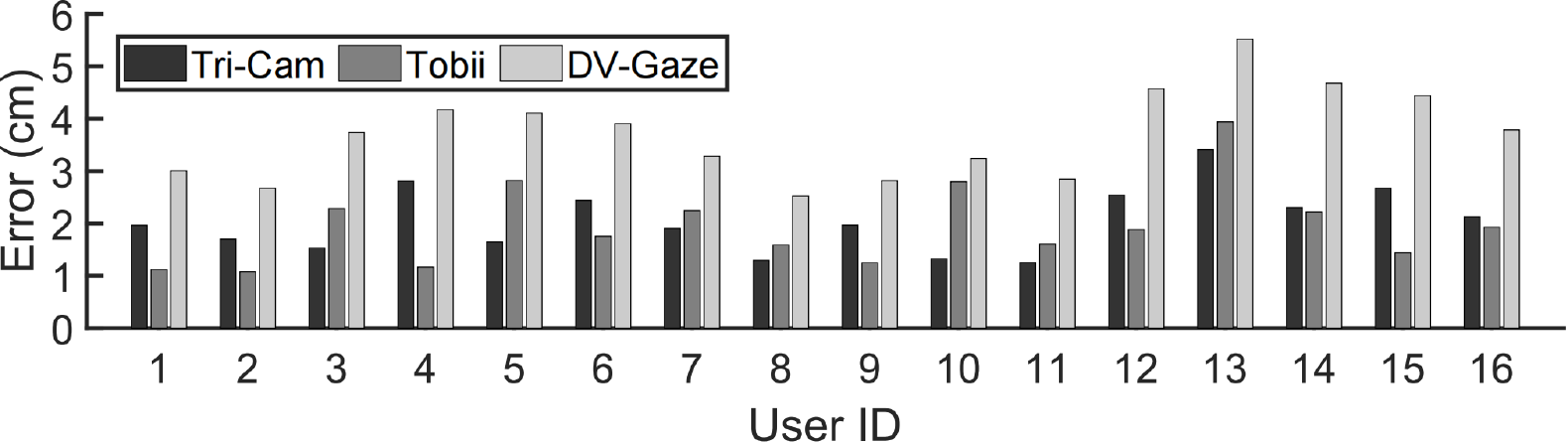}
         \vspace{-5mm}
        \caption{Gaze tracking error on different users}
        \label{fig_expm_users}
    \end{minipage}
    \hfill
    \Description[bruh]{bruh bruh}
\end{figure}

As shown in Figure \ref{fig_expm_users}, Tri-Cam has an overall similar performance to Tobii.
The average gaze estimation error for Tri-Cam, Tobii, and DV-Gaze are 2.0558cm, 1.9455cm, and 3.7077cm, respectively. 
Although Tri-Cam and DV-Gaze share the same webcams as input, DV-Gaze is designed to use two cameras, while Tri-Cam uses three, which potentially provides more visual details.
Additionally, DV-Gaze uses the face image, instead of cropped eye image, which may introduce more visual influence to the convolutional neural networks, e.g., facial landmarks like nose and mouth will also be considered by the neural network, while having little relevance to the gaze inference.
Tri-Cam, on the other hand, focuses only on the cropped eye images, which have concentrated information relevant to gaze tracking, and thus could potentially learn and perform better.

Compared to Tobii, Tri-Cam has an overall similar performance.
For certain users, the error of Tobii eye tracker is a lot higher, which is possibly due to the eyes' biological differences.
Tobii eye tracker uses infrared signal to track the gaze, as human eyes have special interaction with the infrared, which is dependent on the eyes' biological nature.
Therefore, even though it has been calibrated by the user before using, Tobii could still have fluctuating accuracy across different users.
Additionally, for users that wear glasses, their glasses are made of different materials, among which, resin glasses are usually transparent to infrared, while glass-based materials could block the infrared spectrum, making Tobii inaccurate.
On the other hand, Tri-Cam is relatively consistent across users, as we only use the visible light spectrum.

\subsubsection{Different Gaze Angles}
\label{sec_expm_angle}

We also test the accuracy of Tri-Cam at different view angles.
This angle refers to the angle between the user's gaze direction and the perpendicular direction of the monitor screen.
While maintaining a distance of 50cm from the monitor, we set a horizontal offset for the users' seat location, starting at the central position facing the monitor.
Different gaze angle $\theta$ results in different offset $\Delta x$:

\begin{align}
    \theta & \in \{ -30\degree, -20\degree, -10\degree, -5\degree, 0\degree, 5\degree, 10\degree, 20\degree, 30\degree \} \\
     &\Delta x  = L \cdot  tan(\theta) \quad\quad\quad (L=50cm)
\end{align}

\begin{figure}[ht]
    \hfill
    \begin{minipage}[t]{0.99\linewidth}
        \centering
        \includegraphics[width=\textwidth]{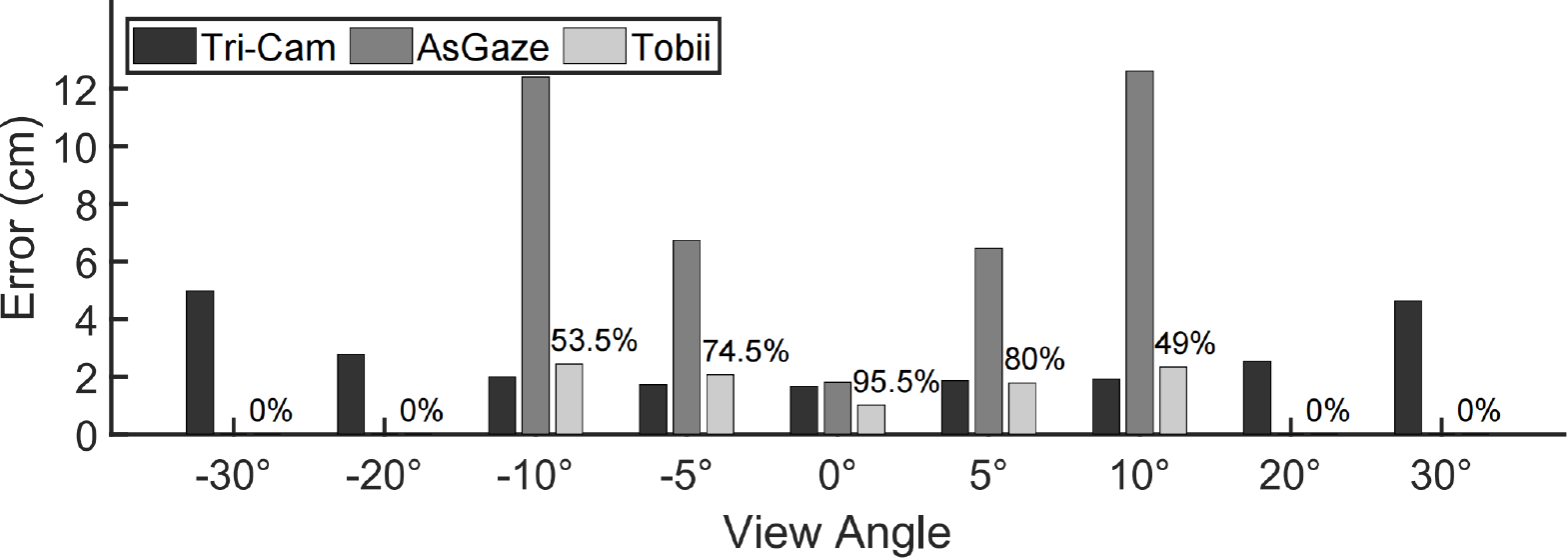}
         \vspace{-5mm}
        \caption{Performance at different view angles}
        \label{fig_expm_angle}
        \vspace{-3mm}
    \end{minipage}
    \hfill
    \hfill
    \Description[bruh]{bruh bruh}
\end{figure}

Figure \ref{fig_expm_angle} shows the average gaze estimation error at different gaze angle, as well as the detection rate of Tobii, denoted on each bar that represents Tobii's performance.
$0 \degree$ means the user sits along the central axis of the monitor, where Tobii is very accurate and robust.
Its error at $0 \degree$ is 
1.0049cm
, and detection rate is 95.5\%.
However, as soon as the user move to $\pm5\degree$, Tobii's detection rate begins to drop, and error begins to rise.
If the absolute gaze angle exceeds $10\degree$, Tobii is unable to detect the eyes and provide gaze inference.
This is highly possibly due to the single-camera configuration of Tobii.
Although the infrared camera of Tobii is powerful at gaze tracking, it could still only cover limited area from a single origin.
AsGaze has an error of 1.8091cm at $0 \degree$, but intensively increases along view angle.
Single-non-depth-camera systems cannot handle free movement.
It is only accurate at its initial position, namely, $0 \degree$, where it was calibrated.

On the other hand, since Tri-Cam utilizes three cameras, it is always able to detect the eyes in the view of at least two of its cameras.
The error of Tri-Cam also rises as the absolute gaze angle increases, but in a slower trend.
We notice that at $\pm30\degree$ view angle, there would be one camera that cannot capture the user in its view, thus Tri-Cam will have to rely on the rest of the two cameras.
Therefore, the error of Tri-Cam largely increases to around 5cm.
However, looking at the screen with a $\pm30\degree$ angle is not comfortable or common, and Tri-Cam is able to maintain a relatively stable performance within the range of [$-20\degree,20\degree$].
This indicates the potential of Tri-Cam to support free-movement gaze tracking.

\subsubsection{Spatial Distribution of Accuracy}

\begin{figure}[ht]
    \hfill
    \begin{minipage}[t]{0.95\linewidth}
        \centering
        \includegraphics[width=\textwidth]{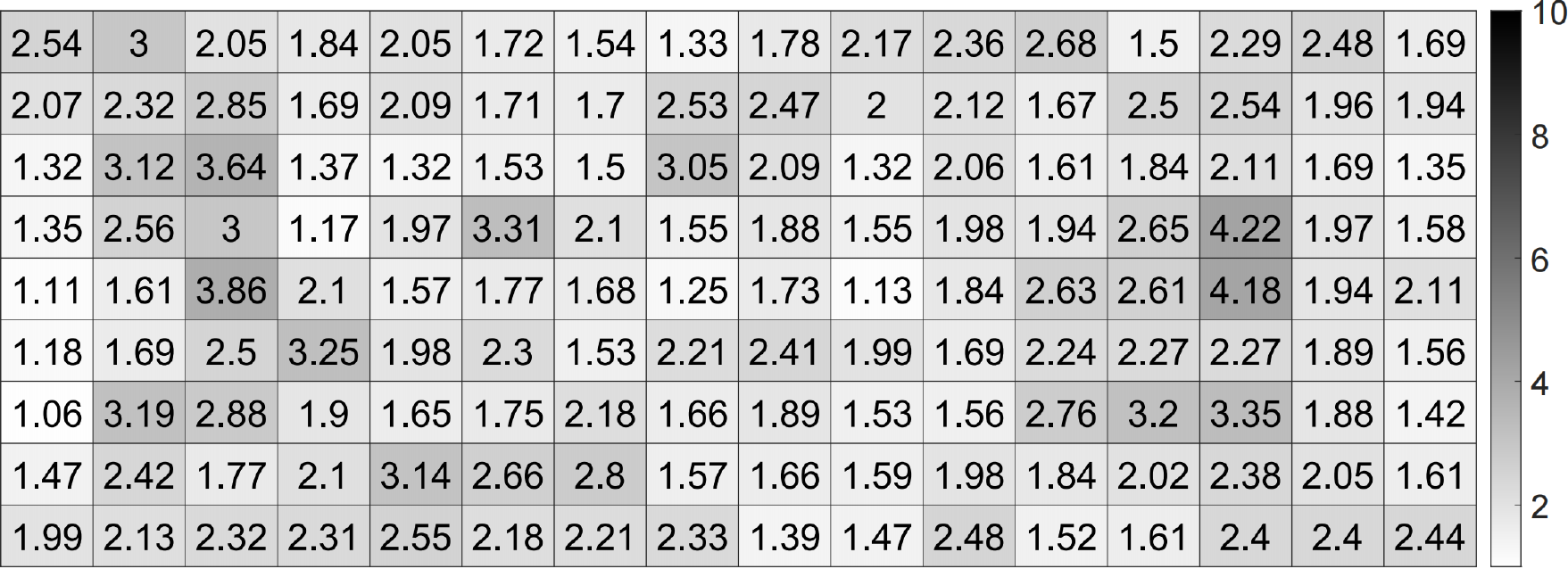}
         \vspace{-5mm}
        \caption{Spatial distribution of gaze tracking error}
        \label{fig_expm_loc}
        \vspace{-3mm}
    \end{minipage}
    \hfill
    \Description[bruh]{bruh bruh}
\end{figure}

Figure \ref{fig_expm_loc} shows the spatial distribution of the gaze estimation error of Tri-Cam.
The overall error distribution is relatively even.
The central area has the overall best accuracy, since the user is directly facing the monitor when looking at the center of it.
The camera network receives maximum image quality at this angle.
The performance on borders and corners are also acceptable, as Tri-Cam utilizes three cameras, each one of which serves as backup for the others, and are usually able to find a valid angle of view to the eyes.

However, there are two specific clusters that have relatively high error, located around $(300,480)$ and $(1600,480)$.
As a possible explanation, for users who wear glasses, looking at these two coordinates will reflect more light from the screen to the camera on either left or right side.
These two areas are slightly above the middle line, since the cameras are deployed on the top of the monitor, and would receive the reflection when the user is looking slightly above. 
Assume the user is looking at the $(1600,480)$ area.
The reflected light will block the eye image and make the right camera unreliable.
Therefore, Tri-Cam has to focus more on the middle and left cameras, which would have an angled view to the user's eye, especially for the left camera, as the user is looking to the right side.
On the other hand, if the user is looking at the center and the reflected light is blocking the vision of the middle camera, both left and right cameras can serve as backup, and have better angle of view, compared to the side situations.
Therefore the middle area does not have a serious reflection issue, which is also possibly due to the compensation of the high image quality at this angle.

\subsubsection{Unseen Users}

\begin{figure}[ht]
    \hfill
    \begin{minipage}[t]{0.99\linewidth}
        \centering
        \includegraphics[width=\textwidth]{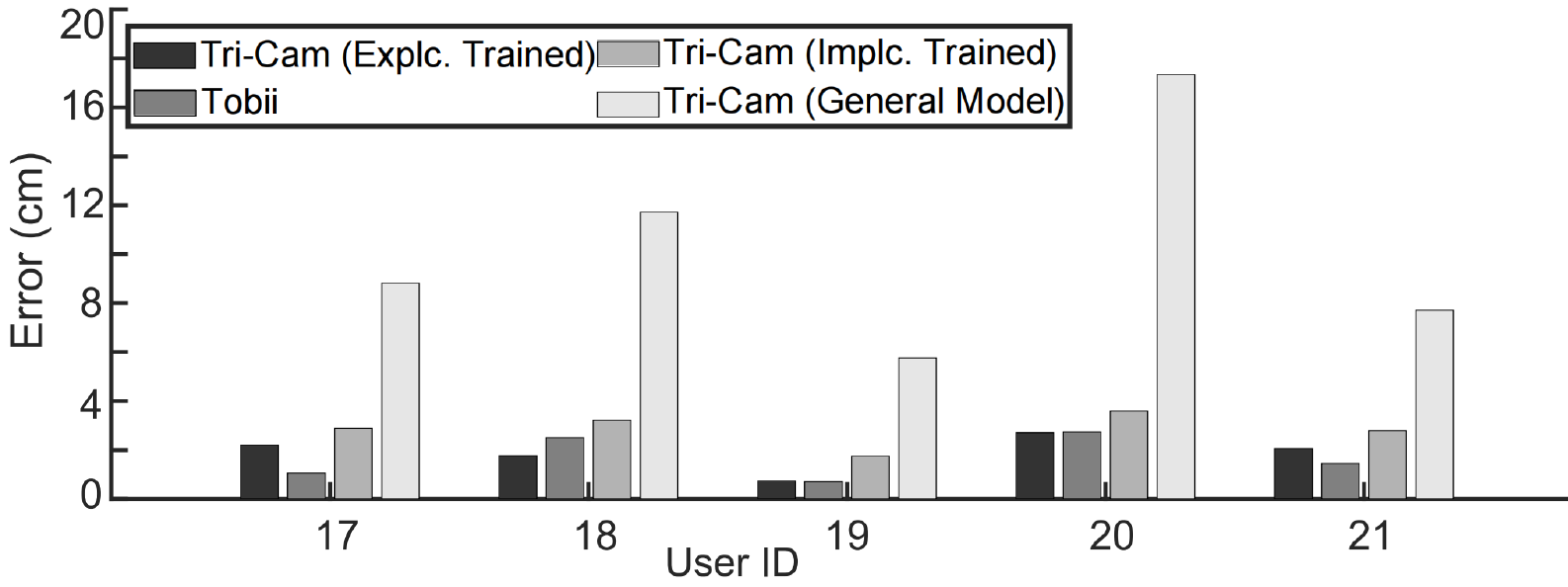}
         \vspace{-5mm}
        \caption{Performance on unseen users}
        \label{fig_expm_unseen}
    \end{minipage}
    \hfill
    \hfill
    \Description[bruh]{bruh bruh}
\end{figure}

To test the scalability of Tri-Cam, we train a general model using all training data from user $\#1\sim16$.
The average error of the general model on the test data of user $\#1\sim16$ is
3.0031cm.
Then, we test the general model on five unseen users ($\#17\sim21$).
As shown in Figure \ref{fig_expm_unseen}, the general model performs very bad on the five unseen users, especially for user $\#20$, who wears glasses.
The average error among the five users is 
10.2674
cm, which is not capable of supporting gaze tracking applications.
If the model is trained with the user's personal training data, it can perform relatively well, and the average error among the five users is 
1.8918
cm.
On the unseen users' data, Tobii still has a stable performance, with an average error of 
1.6864
cm.

This result indicates that Tri-Cam does not perform well without calibration (i.e., being trained with new users' data).
The main reason lies in the difference of appearance among users.
As shown in Figure \ref{fig_user_photo}, the size and shape or the eyes, as well as facial feature around the eyes, vary largely among users.
Additionally, the gaze pattern of users may also be biologically different.
Therefore, it could be difficult for the convolutional neural network to know how to track the gaze of unseen users.
Note that, the Tobii Pro Spark eye tracker is always calibrated with the new user's participation before being used, while this condition is missing for Tri-Cam, as it is not trained with new users' data.
Scalability has been a popular topic for neural network design.
However, we leave the possibility of improving it with further complex design to future works.
Instead, we rely on the implicit re-calibration module to adapt to new users' feature.

\textbf{Implicit re-calibration on unseen user.}
New users can always re-calibrate Tri-Cam via performing explicit data collection and re-train the Tri-Cam model.
It is also viable to use the implicit data collection module and re-calibrate Tri-Cam without having to stare at visual targets.
Therefore we also show the performance of Tri-Cam trained with implicitly collected data from the five unseen users.
From Figure \ref{fig_expm_unseen}, we can see that the error of the implicitly trained model is higher than the explicitly trained model.
This is due to the alignment error of the implicitly collected click samples.
According to the supervision of Tobii, the expectation of the alignment error of the click samples is 3.28cm, after all three criteria have been applied.
With these sub-accurate training data, Tri-Cam reaches an average error of 
2.8087
cm, which is 
48.47\% higher than the explicitly trained model.
This increase is much smaller than the alignment error of 3.28cm, which is mainly due to two reasons.
First, the alignment error is evaluated using Tobii as temporary ground truth, which may not be absolutely accurate.
Second, the difference vector of from the implicitly collected click samples to the actual gaze point during the click could be in various directions, and thus, aggregating them could potentially decrease the overall alignment error.
Therefore, in the batched training of Tri-Cam using the implicitly collected data, the impact of the alignment error could be lower than the absolute value of the alignment error.
In conclusion, the implicit data collection module is capable of training Tri-Cam to a applicable level of accuracy, without having to ask the user to explicitly stare at visual targets to collect gaze data.
This module is also capable of supporting low-effort re-calibration of the model in the case of new monitor-camera setup.

\subsection{Training Efficiency and Overhead}

We now demonstrate the training efficiency of Tri-Cam.
For the 4000 data samples collected by each user, 2800 of them are used for training, and 700 are used for verification.
Timer shows that training with all 2800 training data takes 2.1305 seconds per  one epoch.
On the other hand, DV-Gaze takes 32.2898 seconds per epoch.
This difference is due to the neural network size and complexity.
Compared to the tiny cropped images used by Tri-Cam, DV-Gaze uses the face image as input, which demands larger convolutional neural networks to handle the larger image input.
DV-Gaze also uses both Resnet and Transformer, which are known to be computationally intensive.
From the statistical aspect, Tri-Cam has 1245113 parameters, while DV-Gaze has 13619382 parameters.
Higher complexity results in larger optimization overhead.
Tri-Cam divides the camera-eye geometry and the eye-screen geometry, and uses relatively lightweight neural networks to handle them, resulting in faster training.

\begin{figure}[ht]
    \hfill
    \begin{minipage}[t]{0.42\linewidth}
        \centering
        \includegraphics[width=\textwidth]{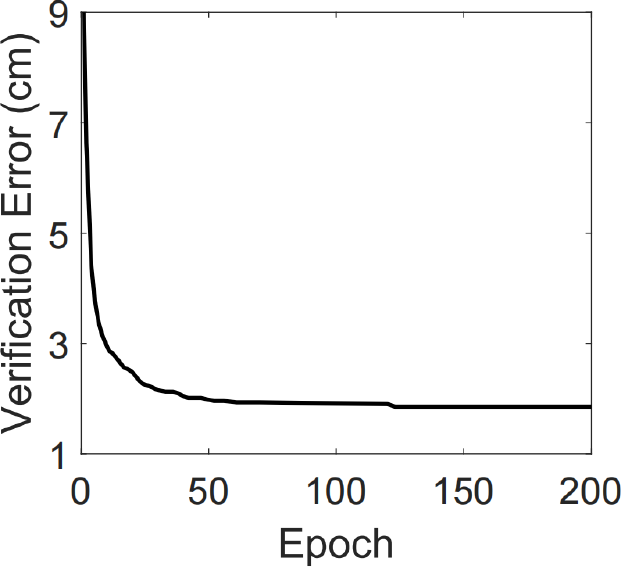}
         \vspace{-5mm}
        \caption{Verification error during training}
        \label{fig_expm_train}
    \end{minipage}
    \hfill
    \begin{minipage}[t]{0.53\linewidth}
        \centering
        \includegraphics[width=\textwidth]{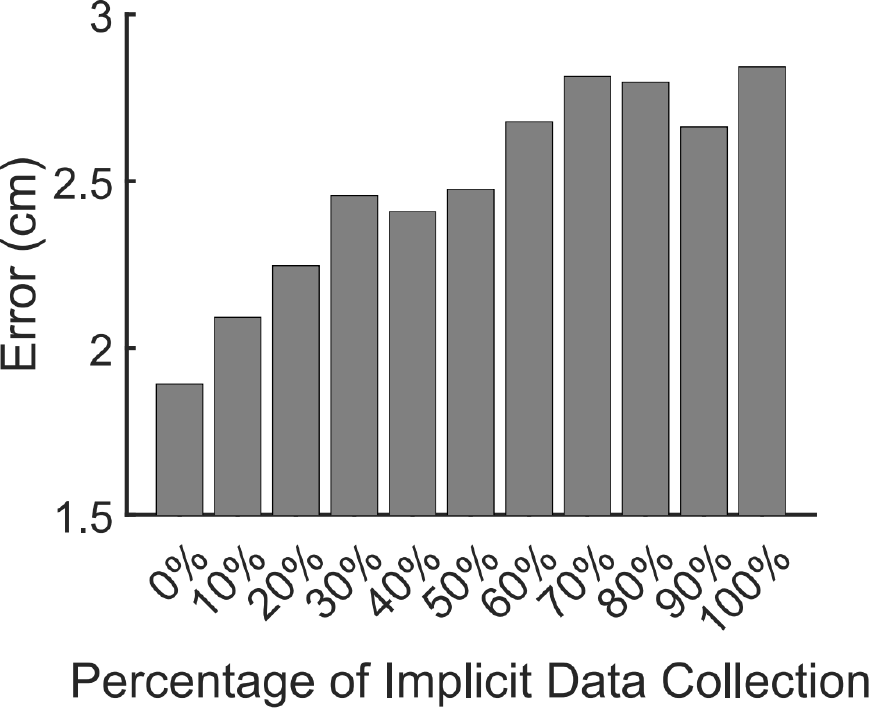}
         \vspace{-5mm}
        \caption{Combining explicit and implicit Data collection}
        \label{fig_expm_implic}
    \end{minipage}
    \hfill
    \Description[bruh]{bruh bruh}
\end{figure}

Figure \ref{fig_expm_train} also shows the model verification error of Tri-Cam during training, averaged among users.
We can see that the model's verification error converges after 150 training epochs, which is within 6 minutes.
This guarantees that the user do not need to wait long for Tri-Cam to be ready to use.


\textbf{Combining Explicit and Implicit Data Collection.}
Beside total explicit data collection and total implicit data collection, it is also viable to combine the implicit and explicit data collection, if the user is not satisfied with the with the implicitly trained model.
This could potentially reduce the overhead of data collection and avoid eye fatigue.
To evaluate the effect of the combination of explicit and implicit data collection, among the 2800 explicitly collected training data samples, we replace certain percentage of them with equal amount of data samples implicitly collected from the same user.
With different percentage setting, we train a individual model, and test its gaze tracking accuracy.
Figure \ref{fig_expm_implic} shows the average error of each model tested on user $\#17\sim \#21$.
We can see as the percentage of implicit data collection increases, the error rises accordingly.
It is because the implicitly collected data has higher alignment error by its nature, i.e., the user's gaze does not completely align with the visual target, while the visual target's location is being used as gaze point ground truth.
We also notice that when at a low percentage (<30\%), the percentage itself has a greater impact on the model accuracy.
As a speculation, small portion of implicitly collected data is enough to effectively pollute the data pool, confusing the model during training.
This implies that it is preferred to either use implicit collection more (>30\%), or not use it at all, when in combination with explicit data collection.


\subsection{Ablation Study}

We examine the contribution of each system component, via cutting it off and test the performance of Tri-Cam.

\begin{figure*}[ht]
    \hfill
    \begin{minipage}[t]{0.22\linewidth}
        \centering
        \includegraphics[width=\textwidth]{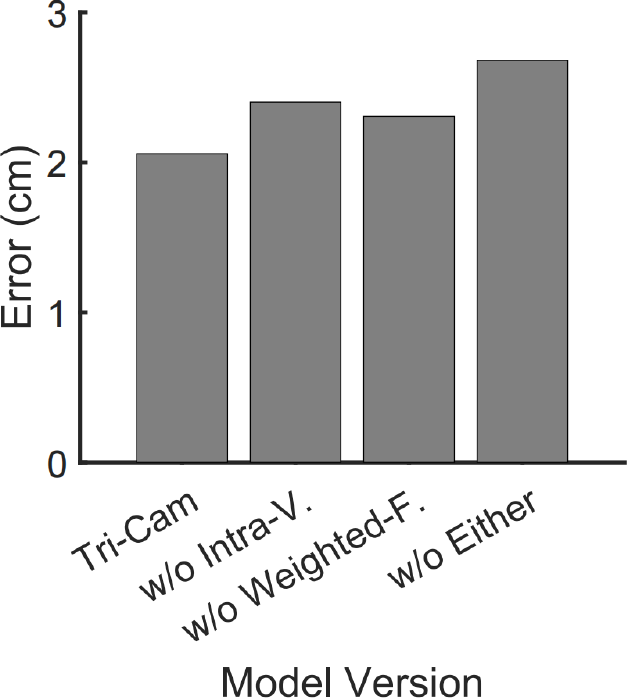}
         \vspace{-5mm}
        \caption{Design component ablation study}
        \label{fig_expm_ablation}
    \end{minipage}
    \hfill
    \begin{minipage}[t]{0.24\linewidth}
        \centering
        \includegraphics[width=\textwidth]{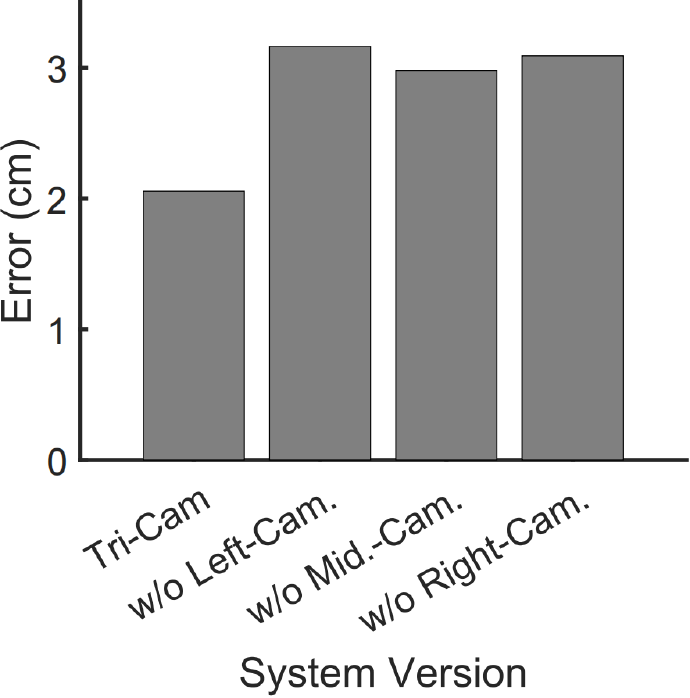}
         \vspace{-5mm}
        \caption{Importance of each camera}
        \label{fig_expm_cut_camera}
    \end{minipage}
    \hfill
    \begin{minipage}[t]{0.18\linewidth}
        \centering
        \includegraphics[width=\textwidth]{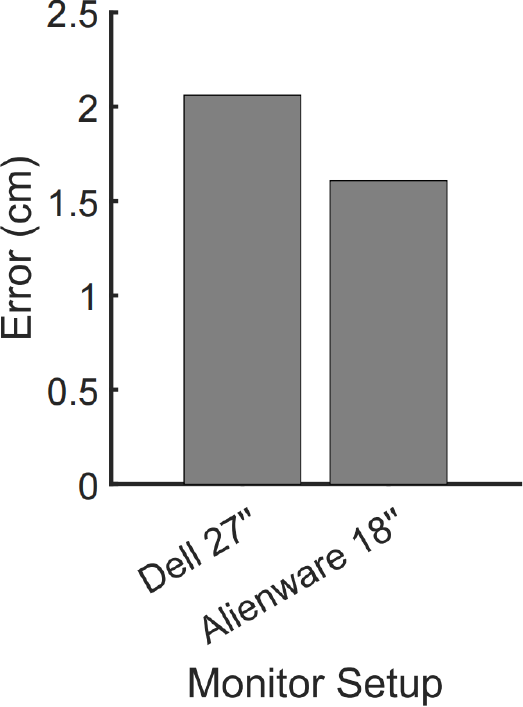}
         \vspace{-5mm}
        \caption{Different monitor setups}
        \label{fig_expm_monitor}
    \end{minipage}
    \hfill
    \begin{minipage}[t]{0.26\linewidth}
        \centering
        \includegraphics[width=\textwidth]{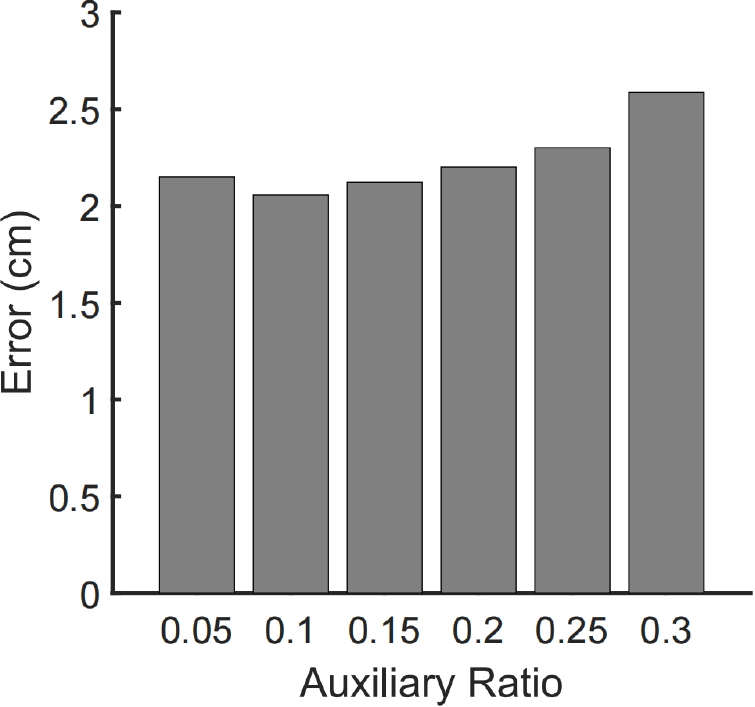}
         \vspace{-5mm}
        \caption{Influence of auxiliary ratio}
        \label{fig_expm_aux_ratio}
    \end{minipage}
    \hfill
    \hfill
    \Description[bruh]{bruh bruh}
\end{figure*}

\subsubsection{Cutting off Intra-validation}

We first cut off the intra-validation component, and the gaze tracking error of Tri-Cam rises to 2.4015cm, which is 16.82\% higher than the complete model.
It shows that the auxiliary multitasking is capable of supporting the training of the main model, enabling extra back propagation via the auxiliary outputs.
The intra-validation design does not introduce any other information or measurement into the system, but instead, exploits the intra-relationship between cameras' view angle to improve the performance of Tri-Cam.

\subsubsection{Cutting off Weighted Fusion}

Second, cutting off the weighted-fusion design increases the error of Tri-Cam to 2.3065cm, which is 12.19\% higher.
This indicates that the discriminator is capable of judging the cropped eye images before they are considered in the final decision layer, and granting weights to the images, so that the neural network can focus less on images with lower quality.

\subsubsection{Cutting off One Camera}

Last, to examine the importance of each camera, we cut off one camera at a time.
As shown in Figure \ref{fig_expm_cut_camera}, all three cameras are important to Tri-Cam.
Without any of three cameras, the error rises to around 3cm, due to the missing support from the intra-validation mechanism.
Cutting off one of the side cameras appears to be slightly more deadly to Tri-Cam than cutting off the middle camera.
The distance between the side cameras is twice the distance between the middle camera and either of the side cameras, which could potentially support better depth perception, and thus makes the version without the middle camera slightly better than the others.

\subsection{Other Factors of Interest}


\subsubsection{Different Monitor-Camera Setup}

We also setup the webcams on the Alienware laptop and use its 18'' screen as a second monitor for experiment.
User $\#21$ collected data on this laptop, and thus we only compare the results of user $\#21$ on the two monitor setups.
As shown in Figure \ref{fig_expm_monitor}, the gaze tracking error is 2.0610cm and 1.6075cm for the 27'' and 18'' monitor setups, respectively.
First, it demonstrates that Tri-Cam can be applied on different monitor setups.
Second, the decrease in the gaze tracking error is probably due to the confined space of possible gaze point.
Specifically, the area of the $1920\times1080$-27'' screen is 2010.76 $cm^2$, while the $1920\times1200$-18'' screen is 939.74 $cm^2$, which is approximately half of the 27'' screen's area.
It is similar to cutting a half from the 27'' screen, and only use the gaze data within this area for training and testing, which makes the task simpler.

\subsubsection{Inference Latency}

Statistics show when running on the Alienware laptop, Tri-Cam takes 0.1741ms to infer one gaze point, which ensures that the maximum real time inference rate of Tri-Cam is 5745Hz.
Although other platforms may have less computation power, this result is still more than enough to support the 30Hz webcams as input.
This is mainly due to the lightweight network design of Tri-Cam.
In potential real application, it is very unlike that the network inference time of Tri-Cam would be the bottleneck, when collaborating with other components, such as webcam recording, eye detection algorithm, gaze result rendering, etc..

\subsubsection{Auxiliary Ratio}

We set 0.1 as the auxiliary multitasking ratio, which will be multiplied onto the auxiliary loss before it is incorporated into the joint loss.
We also test the performance of Tri-Cam with different choices of the auxiliary ratio.
Figure \ref{fig_expm_aux_ratio} shows the gaze tracking error of Tri-Cam with different auxiliary ratio settings.
We can see that 0.1 provides the best accuracy.
Lower auxiliary ratio will eliminate the effect of intra-validation via auxiliary multitasking, while higher auxiliary ratio will over stress the intra-validation, particularly given that there are six auxiliary outputs.

    \section{Conclusion}

This paper introduces Tri-Cam, a practical deep learning-based gaze tracking system using three affordable RGB webcams, featuring a split network structure for efficient training, as well as designated network designs to handle the separated gaze tracking tasks.
An implicit calibration module is deployed to make use of mouse click opportunities to reduce calibration effort for the user. 
Experiment shows Tri-Cam achieves comparable accuracy to Tobii, the state-of-the-art commercial eye tracker, while supporting a wider free movement area. 
In conclusion, Tri-Cam provides a user-friendly, affordable, and robust gaze tracking solution that could practically enable various applications.

\end{document}